\documentclass[sigconf]{acmart}

\AtBeginDocument{%
  \providecommand\BibTeX{{%
    \normalfont B\kern-0.5em{\scshape i\kern-0.25em b}\kern-0.8em\TeX}}}

\setcopyright{none}
\renewcommand\footnotetextcopyrightpermission[1]{} 
\begin{document}

\title{Neuroevolutionary Multi-objective approaches to Trajectory Prediction in Autonomous Vehicles}



\author{Fergal Stapleton}
\authornote{Joint first/lead authors}
\email{fergal.stapleton.2020@mumail.ie}
\affiliation{%
  \institution{Dept. of CS, Hamilton Institute, Maynooth University, \\ Naturally Inspired Computation Res. Group}
 \country{Ireland}
}

\author{Edgar Galv\'an}
\authornotemark[1]
\email{  edgar.galvan@mu.ie}
\affiliation{%
  \institution{Dept. of CS, Hamilton Institute, IVI, Maynooth University, \\ Naturally Inspired Computation Res. Group}
  \country{Ireland}
}



\author{Ganesh Sistu}
\email{ganesh.sistu@valeo.com}
\affiliation{%
  \institution{Valeo Vision Systems,}
  \state{Galway}
  \country{Ireland}
}

\author{Senthil Yogamani}
\email{senthil.yogamani@valeo.com}
\affiliation{%
  \institution{Valeo Vision Systems,}
  \state{Galway}
  \country{Ireland}
}






\setlength{\textfloatsep}{2.0pt plus 2.0pt minus 2.0pt}

\begin{abstract}


The incentive for using Evolutionary Algorithms (EAs) for the automated optimization and training of deep neural networks (DNNs), a process referred to as neuroevolution, has gained momentum in recent years. The configuration and training of these networks can be posed as optimization problems. Indeed, most of the recent works on neuroevolution have focused their attention on single-objective optimization. Moreover, from the little research that has been done at the intersection of neuroevolution and evolutionary multi-objective optimization (EMO), all the research that has been carried out has focused predominantly on the use of one type of DNN: convolutional neural networks (CNNs), using well-established standard benchmark problems such as MNIST. In this work, we make a leap in the understanding of these two areas (neuroevolution and EMO), regarded in this work as neuroevolutionary multi-objective, by using and studying a rich DNN composed of a CNN and Long-short Term Memory network. Moreover, we use a robust and challenging vehicle trajectory prediction problem. By using the well-known Non-dominated Sorting Genetic Algorithm-II, we study the effects of five different objectives, tested in categories of three, allowing us to show how these objectives have either a positive or detrimental effect in neuroevolution for trajectory prediction in autonomous vehicles.


\end{abstract}


\maketitle

\pagestyle{plain}

\section{Introduction}
\label{sec:intro}

Predicting future trajectories in autonomous vehicles has the potential to produce safer driving environments for road users while alleviating the need for human interaction, which can be prone to error. As such, the automated trajectory of vehicles is a key area of research in autonomous driving and multiple works have emerged in recent years~\cite{buhet2020plop,Mo2020InteractionAwareTP,Xie2021CongestionawareMT}. It remains a challenging problem in autonomous driving compared to other perception tasks where deep learning models have performed exceptionally well~\cite{chennupati2019auxnet, kumar2021svdistnet, kumar2020syndistnet, kumar2021omnidet, kumar2020unrectdepthnet}.


Deep Neural Networks (DNNs)~\cite{LeCun2015} can be effective machine learning techniques to generate perception models for planning, applied in different areas including in facial recognition~\cite{Galvan_Face_GECCO_2022} and studied from different perspectives~\cite{DBLP:journals/corr/abs-2102-08475}. The implementation of these models pose their own unique set of challenges, such as finding the correct set of hyperparameters’ values for training the network, which is the focus of this work. This challenge is compounded when multiple objectives, in conflict or not, are considered. One can naturally tackle this using evolutionary multi-objective optimization (EMO) ~\cite{CoelloCoello1999,1597059}. However, as articulated in a recent IEEE Trans. on AI article on neuroevolution in deep neural networks by Galv\'an and Mooney, covering over 170 recent works on neuroevolution~\cite{9383028}, little research has been done at the intersection of these two areas: neuroevolution and EMO, where all the research carried out by the research community has focused predominately on the use of one type of Deep Learning (DL) network.  




The first contribution of this work is to use a rich DL network suitable for this task, as such, we use a network composed of a Convolutional Neural Network (CNN)~\cite{lecun1998gradient} and Long-Short Term Memory (LSTM) network~\cite{10.1162/neco.1997.9.8.1735}, using a larger hyperparameter search space than previously reported in multi-objective (MO) trajectory prediction~\cite{neurotrajCode,grigorescu2019ieee}. The second contribution of this work is to shed light on the type of objectives that might be beneficial or detrimental in neuroevolution for trajectory prediction, as well as sharing insight into the conflicting nature of these objectives with respect to each other. The third contribution is to use a well-established EMO approach, the Non-dominated Sorting Genetic Algorithm II (NSGA-II)~\cite{996017} to test and validate our approach, contrary to the works carried out in this area of autonomous vehicles that have limited their attention on the use of a more restrictive EMO approach~\cite{grigorescu2019ieee}. These contributions will highlight the importance of certain objectives for the correct trajectory prediction in autonomous vehicles.

\section{Methodology}
\label{sec:methodology}
\subsection{Trajectory Prediction and Objective Optimization}

We need input data to effectively carry out trajectory prediction. The input data consists of sequenced occupancy grids captured using the GridSim simulator~\cite{trasnea2019gridsim}. We can define a sequence as consisting of $\tau$ number of images (or occupancy grids) as $X^{<t-\tau>}$ at time $t$. Using $X^{<t-\tau>}$, the aim is to predict future trajectory positions 

\begin{equation}
\label{eqn:position}
Y^{<t+\tau>} = \{x_0, y_0,\  x_1, y_1, \  x_2, y_2,\  ... , \ x_\tau , y_\tau\}
\end{equation}

\noindent where $x$ and $y$ represent the position of the ego vehicle. 




The distance feedback $l_1^{<t+\tau>}$ measures the distance between the current position of the ego vehicle $P_{ego}$ and last position in the sequence $P_{dest}$ and is expressed in Equation~\ref{eqn:l1}. The aim of this objective is to reduce the local travel path of the ego vehicle.
\begin{equation}
    \label{eqn:l1}
   l_1^{<t+\tau>} = \sum\limits_{i=1}^{\tau_0}\vert\vert P_{ego}^{<t+i>} - P_{dest}^{<t+\tau>}\vert\vert_{2}^{2}
\end{equation}

The lateral velocity $l_2$ is calculated from the angular velocity of the ego vehicle $v_{\delta}$, as seen in Equations~\ref{eqn:l2}. The aim is to reduce sudden or rapid movements.

\begin{equation}
    \label{eqn:l2}
   l_2^{<t+\tau>} = \sum\limits_{i=1}^{\tau_0} v_{\delta}^{<t+i>} 
 \end{equation}  

The longitudinal velocity $l_3^{<t+\tau>}$ is calculated as the component of the velocity in the $y$-direction, as defined in Equation~\ref{eqn:l3}. Lower and upper bounds have been set on the velocity of 80km for $v_{min}$ and 130km for $v_{max}$. The aim of this objective is to help shorten the overall travel time for the passenger.



\begin{equation}
    \label{eqn:l3}
   l_3^{<t+\tau>} = \sum\limits_{i=1}^{\tau_0} v_{f}^{<t+i>} \in [v_{min}, v_{max} ]
\end{equation}

The root mean squared error ($RMSE$) was also tested as an objective and is given as the Euclidean distance between the predicted position $\hat{P}_{ego}$ of the ego vehicle for a given trajectory at a given time-step and the actual position of the ego vehicle $P_{ego}$ at that time step as seen in Equation~\ref{eqn:rmse}. 


\begin{equation}
    \label{eqn:rmse}
   RMSE = \sum_{i=1}^n \frac{\sqrt{(\hat{P}_{ego} - P_{ego})^2}}{n}
\end{equation}

Another consideration, when considering the highway data, is when a poor model incorrectly predicts that the trajectory should always veer to either in the left or right direction. Using the trajectory data, pre-scaled, it is possible to incorporate the sign as an indication of the direction (negative $x$ for left and positive $x$ for right). Furthermore, it is possible to penalize networks which calculate the ego vehicle to go straight all the time by taking the absolute value of the predicted away from the real trajectory. These two criteria are combined in Equation~\ref{eqn:sign_loss}:

\begin{equation}
    \label{eqn:sign_loss}
   Sign Loss = \frac{\sum_{i=1}^n \frac{\sqrt{(\|\hat{X}_{ego}\| - \|X_{ego}\|)^2}}{n} }{I(sign(\hat{X}) == sign(X))}
\end{equation}



\subsection{Network Topology}

\begin{figure}
  \centering
  \includegraphics[width=0.9\linewidth]{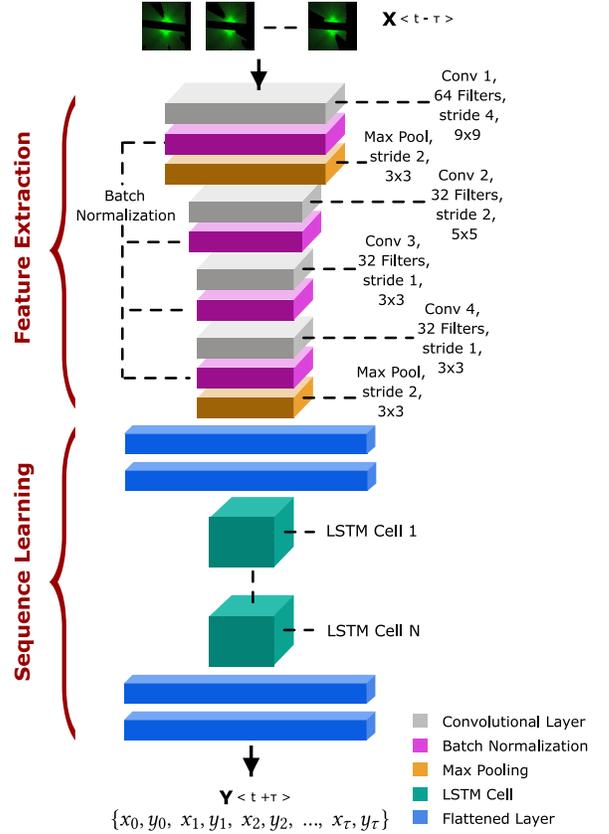}
  \caption{Diagram of network topology. On the top of the diagram, sequences of image of size $\tau$ are fed as input into the CNN. On the bottom is the outputted predicted trajectory.}
  \label{fig:network}
\end{figure}

Broadly speaking, the network is composed of two parts: (i) the Convolutional Neural Network (CNN) which takes the sequence of $\tau$ images as input and which is responsible for extracting important feature information from these images and (ii) the Long-Short Term Memory Network (LSTM) which predicts the trajectories based on the output of the CNN. LSTMs are particularly well suited for temporal prediction-based problems using sequenced input data. Since each position represents the location of the ego vehicle at future point in time, the aim is for our LSTM network to learn these future positions. A summary of the architecture topology is shown in Figure~\ref{fig:network}. A number of fixed hyperparameters are selected, primarily for the CNN section of the network, and these fixed parameters are derived from~\cite{neurotrajCode}. Just to note, for more complex temporal problems it is often more suitable to chain LSTM blocks, as such the LSTM Cells gene represents a variable number of chained LSTM cells up to size 4, denoted by LSTM Cell N in Figure~\ref{fig:network}.  Table~\ref{tab:hyperparameters} lists the evolvable hyperparameters for each network.

\begin{table}
  \caption{Evolvable hyperparameters for deep neural network.}
  \label{tab:hyperparameters}
  \resizebox{\columnwidth}{!}{ 
  \begin{tabular}{ccc}
    \toprule
    Locus & Gene & Set of possible alleles\\
    \midrule
    1 & Batch Size & \{\ 50, 75, 100, 125\ \} \\
    2 & Epochs & \{\ 10, 20, 30, 40, 50\ \} \\
    3 & Momentum & \{\ 0.8, 0.85, 0.9, 0.95\ \} \\
    4 & Loss Function & \{\ MSE, Log Cosh\ \} \\
    5 & Optimizer & \{\ RMSprop, NAdam, SGD,  \\ &&  AdaGrad, Adadelta, Adam, AdaMax\ \} \\
    6 & LSTM Cells & \{\ 1, 2, 3, 4\ \} \\
    7 & LSTM Dropout & \{\ 0.2, 0.25, 0.3, 0.35, 0.4, 0.5\ \} \\
    8 & Hidden Units & \{\ 100, 125, 150, 175, 200, 225, 250\ \} \\
    9 & CNN Flattened 1 & \{\ 256, 512, 768, 1024\ \} \\
    10 & CNN Flattened 2 & \{\ 256, 512, 768, 1024\ \} \\
    11 & LSTM Flattened 1 & \{\ 64, 128, 256, 512\ \} \\
    12 & LSTM  Flattened 2 & \{\ 64, 128, 256, 512\ \} \\
    13 & Flattened Dropout & \{\ 0.05, 0.1, 0.15, 0.2, 0.25\ \} \\
  \bottomrule
\end{tabular}
}
\end{table}

\section{Experimental Setup}
\label{sec:experimental}

 Over 30km of image data was captured from a highway scenario using the GridSim simulator~\cite{trasnea2019gridsim}. Each network model is trained on a sequence of images, consisting of $\tau$ images of 128 x 128 pixels with three channels for RGB color. The data is split with 0.6 : 0.2: 0.2 ratio between training, validation and test, respectively. For the training set, there are 1500 sequences, and for both the validation and test, 500 sequences each. To increase the sample size, a sliding window approach was used to increase the number of sequences. Additionally, both the training and validation sets were shuffled prior to being split. When setting up our approach we used results from Grigorescu et. al.~\cite{grigorescu2019ieee} as a baseline to ensure our approach was comparable to state-of-the-art results. The test data set was withheld for both training and optimization.
 
 
 
 
 All experiments were run on Nvidia Tesla V100 GPUs for 60 independent runs: 12 runs for each of the five experiments, requiring 25 GPU days. Notice that in the deep learning community, one run is the norm. A further summary of the parameters used for NSGA-II have been documented in Table~\ref{tab:parameters}. A summary of the objective combinations, as discussed in Table~\ref{tab:experimental}, combine an assortment of objectives including distance feedback, lateral velocity, longitudinal velocity, $RMSE$ and $Sign Loss$ (Equation~\ref{eqn:l1} - Equation~\ref{eqn:sign_loss}) for each experiment.

\begin{table}
\centering
\caption{Summary of parameters.}
\resizebox{0.8\columnwidth}{!}{ 
\small\begin{tabular}{|l|r|} \hline 
\emph{Parameter} &
\emph{Value} \\ \hline \hline
Population Size & 25 \\ \hline
Generations & 20 \\ \hline
Type of Cross. and Mut. & Single point  \\ \hline
 Crossover Rate  & 1.00  \\ \hline
Mutation Rate & 0.50 \\ \hline
Selection & Tournament (size = 3)\\ \hline
Total Independent Runs & 60 \\ \hline
\end{tabular}
}
\label{tab:parameters}
\end{table}

\begin{table}
\caption{Objective combinations for each experiment. We have separated the combinations into three categories}.
\begin{tabular}{l|rr}
\toprule

Code & Objectives & Category \\
\midrule
 
 Experiment 1 & $RMSE$, $l_2$, $l_3$  & 1\\
 Experiment 2 & $Sign Loss$, $l_2$, $l_3$ & 1\\
 \hline
 Experiment 3 & $RMSE$, $l_1$, $l_3$  & 2\\
 Experiment 4 & $Sign Loss$, $l_1$, $l_3$ & 2\\
 \hline
 Experiment 5 & $l_1$, $l_2$, $l_3$ & 3\\
\bottomrule
\end{tabular}
\label{tab:experimental}
\end{table}

\begin{table}
\caption{Analysis of trajectory spread denoting `good' models in the final generation of each run.}
\centering
\resizebox{0.48\textwidth}{!}{
\begin{tabular}{rrrrr}
\toprule

Experiment 1 & 
Experiment 2 &
Experiment 3 & 
Experiment 4 &
Experiment 5 \\
\midrule
 

  168/300 & 132/300 & 26/300 & 28/300 & 5/300 \\
  
\bottomrule
\end{tabular}}
\label{tab:spread}
\end{table}

\begin{table*}[t!]
\caption{Metric mean and standard deviation (std) for all runs of a given experiment. Experiments 1 and 2 contain weakly conflicting domain-specific objectives, whereas Experiment 3, 4 and 5 contain strongly conflicting domain-specific objectives. }
\centering
\resizebox{0.90\textwidth}{!}{ 
\begin{tabular}{l|rr|rr|rr|rr|rr}
\toprule

& \multicolumn{2}{c}{Experiment 1} 
& \multicolumn{2}{c}{Experiment 2} 
& \multicolumn{2}{c}{Experiment 3} 
& \multicolumn{2}{c}{Experiment 4} 
& \multicolumn{2}{c}{Experiment 5} \\
 metric &     mean &     $\pm$  std &    mean &     $\pm$ std &     mean &     $\pm$ std &     mean &     $\pm$ std &     mean &     $\pm$ std \\
\midrule

    RMSE$_{val}$ & 0.657413 & 0.032726 & 0.727919 & 0.057119 &  2.077965 & 0.351699 &  2.307763 & 0.234958 &  2.403594 & 0.201088 \\
    RMSE$_{test}$ & 1.063261 & 0.007949 & 1.049083 & 0.010070 &  1.605857 & 0.204658 &  1.734267 & 0.152781 &  1.746888 & 0.131945 \\
\bottomrule
\end{tabular}
}
\label{tab:results}
\end{table*}

\begin{table}
\caption{Spearman rank-order correlation for  distance feedback $l_1$, lateral velocity $l_2$, and longitudinal velocity $l_3$.}
\begin{tabular}{l|rr}
\toprule

Objectives & Coefficient & P - value \\
\midrule
 
$l_1$ \& $l_2$ & 0.1526 & 2.8454e-09 \\
$l_1$ \& $l_3$ & -0.8715 & 0.0 \\
$l_2$ \& $l_3$ & -0.1430 & 2.6771e-08 \\
\bottomrule
\end{tabular}
\label{tab:spearmanr}
\end{table}

\section{ Summary of Results }
\label{sec:results}

The work undertook in this research set out to investigate, whether the inclusion of certain objectives could be beneficial or detrimental for the predictive capabilities of a neuroevolutionary approach to trajectory prediction. To aid our analysis we test that the trajectories had a reasonable spread over the $\tau$ trajectory positions. In other words, trajectories that consistently veer in one location, that fall short in terms of distance travelled or fail to ever change lane can be considered poor models. Table~\ref{tab:spread} denotes models that were deemed to have a `good' spread in terms of the predicted trajectories.

Our findings have demonstrated that the distance feedback function $l_1$, which aims to minimize localized travel path of the ego vehicle, is particularly detrimental when included as an objective. It was found that in the three experiments it was present (Experiments 3 - 5, see Table~\ref{tab:spread}), no experiment was capable of finding any more than 28 useful models out of 300. In the absence of a loss function as an objective, it fared even worse, only finding 5 out of 300 models. Furthermore, the RMSE$_{val}$ and RMSE$_{test}$ results were higher  for these three experiments (Columns 4 - 6, read left to right in Table~\ref{tab:results}) compared to the Experiments 1 and 2 which did not include feedback function $l_1$ (columns 2 - 3, Table~\ref{tab:results}), strengthening our analysis that the inclusion of the distance feedback function was detrimental.

This research highlights the need for the careful consideration of objectives to use in the context of neuroevolution in EMO, with a focus on trajectory prediction in autonomous vehicles. Our analysis showed that the distance feedback function $l_1$ was not conflicting with the lateral velocity $l_2$ and was highly conflicted with the longitudinal velocity $l_3$ (see Table~\ref{tab:spearmanr}). A priori tests for correlation, as such, may be a useful tool for research practitioners, especially when the computational costs of running experiments are considered.


A loss function $Sign Loss$, designed specifically to reduce the error in the $x$-direction by incorporating the direction of the ego vehicle in terms of its sign, was also tested. Notably, as a non-differentiable function it is unsuitable as a loss function for a gradient based neural network, as such it may be of interest to see what effect its inclusion in the EMO may have. With the inclusion of the weakly conflicting objectives it was found that the $Sign Loss$ may help to reduce over-fitting on average (Experiment 2 versus Experiment 1). However, in the presence of highly conflicting objectives the $Sign Loss$ performed worse than $RMSE$.

\section{Conclusion}
\label{sec:conclusions}

Using a robust and well-established EMO approach known as the Non-dominated Sorting Genetic Algorithm II (NSGA-II) we evolved the hyperparameters of deep learning networks, where each network 
was composed of a CNN network and LSTM network. Each network was tasked with predicting vehicle trajectories, a challenging problem domain of high relevance to the field of autonomous vehicles, EMO and DNNs. By analyzing the results of multiple combinations of network-specific and domain-specific objectives, we demonstrated how the inclusion of some objectives can be either detrimental or beneficial to the neuroevolutionary process.  In particular the use of a distance feedback objective was particularly detrimental to the EMO optimizer finding meaningful or useful models. On the other hand, the lateral velocity objective was found to be beneficial in finding meaningful models. When no loss function was present as an objective (i.e when only domain-specific objective were considered), the EMO approach failed in the vast majority of cases to find any meaningful or useful models. A non-differentiable $Sign Loss$ objective was also explored, and it was found that in the presence of the weakly conflicting objectives that were tested (lateral and longitudinal velocity), it helped to alleviate over-fitting on average, suggesting non-differentiable objectives may be worth exploring in future studies in EMO. 




\begin{acks}
        This publication has emanated from research conducted with the financial support of Science Foundation Ireland under Grant number 18/CRT/6049. The authors wish to acknowledge the Irish Centre for High-End Computing (ICHEC) for the provision of computational facilities and support. 
\end{acks}



\bibliographystyle{ACM-Reference-Format}
\bibliography{ref}


\end{document}